\documentclass{article}

 \usepackage[sglblindworkshop, final]{neurips_2026}
\workshoptitle{AI at Scale for Clinical Impact}

\usepackage[utf8]{inputenc} 
\usepackage[T1]{fontenc}    
\usepackage{hyperref}       
\usepackage{url}            
\usepackage{booktabs}       
\usepackage{amsfonts}       
\usepackage{nicefrac}       
\usepackage{microtype}      
\usepackage{xcolor}         
\usepackage{graphicx}
\usepackage{amsmath}
\usepackage{subcaption}
\usepackage{siunitx}

\title{Predicting Single-Cell Spatial Gene Expression from H\&E using Pathology Foundation Models}

\author{%
  \bfseries
  Ashley P. Tsang$^{1,2,*}$ \enspace
  Evan Liu$^{3}$ \enspace
  Conrad Foo$^{4}$ \enspace
  Bob Chen$^{1}$ \enspace
  Hassaan Maan$^{1}$ \enspace
  Donny Chan$^{1}$\\
  \bfseries
  Elisa Penna$^{4}$ \enspace
  Veronica Ibarra Lopez$^{4}$ \enspace
  Hector Corrada Bravo$^{1}$ \enspace
  A\"{i}cha BenTaieb$^{1}$\\
  \bfseries
  Lisa M. McGinnis$^{4}$ \enspace
  Runmin Wei$^{1,\dagger}$ \enspace
  Kenneth T. Gao$^{1,\dagger}$\\[-1em]
  \AND
  \normalfont
  $^{1}$AI for Biology \& Translation, Computational Sciences Center of Excellence, Genentech\\
  $^{2}$Gilbert S. Omenn Department of Computational Medicine \& Bioinformatics, University of Michigan\\
  $^{3}$Computational Biology \& Medicine, Computational Sciences Center of Excellence, Genentech\\
  $^{4}$Digital and Spatial Pathology, Translational Genomics and Proteomics, gRED, Genentech\\
  \footnotesize
  $^{*}$Work conducted during an internship at Genentech.\quad
  $^{\dagger}$Corresponding author.
}

\begin{document}
\maketitle
\begin{abstract}
Predicting gene expression directly from hematoxylin and eosin (H\&E) images offers a scalable and cost-effective approach to deriving spatial molecular insights from routine histology. Pathology foundation models, typically pretrained using tile-level objectives, may not adequately capture the fine-grained cellular features required for single-cell molecular prediction. In this work, we systematically evaluate and establish domain-informed and task-specific strategies for adapting pathology foundation models to single-cell spatial gene expression prediction. Using paired H\&E and Xenium 5K data, our approach improves mean gene-wise Pearson correlation by 2.67$\times$ over the standard zero-shot baseline and achieves state-of-the-art performance on this task. We demonstrate strong gene-detection performance across the full gene panel (AUROC = 0.800) and importantly identify that predictive performance varies substantially across biological contexts, with our model achieving its highest Pearson correlation for spatially variable genes ($r$ = 0.643). Our findings demonstrate that, despite the challenges of sparse single-cell measurements, H\&E-derived representations can recover biologically meaningful spatial expression patterns, highlighting both the opportunities and limitations of morphology-based molecular inference at cellular resolution.
\end{abstract}
\section{Introduction}
\label{sec:intro}

Spatial transcriptomics (ST) enables the study of tissue biology by measuring gene expression in situ~\cite{rao2021exploring, staahl2016visualization, moncada2020integrating}. However, these assays are costly, experimentally demanding, and limited in the number of samples that can be profiled at scale. In contrast, hematoxylin and eosin (H\&E) staining is inexpensive, routinely collected, and widely available in clinical archives. H\&E images capture cellular morphology and tissue organization that reflect underlying molecular states and provide visual signals informative of spatial gene expression~\cite{stnet, he2rna, levy2020spatial}. Thus, predicting gene expression from H\&E offers a scalable, cost-effective approach to spatial molecular profiling.

Existing H\&E-to-ST methods have been developed predominantly for spot-level ST data~\cite{stnet, hist2st, bleep, omiclip, wang2025benchmarking}. In these settings, gene expression is measured at spatially arrayed capture spots, each of which aggregates transcripts from multiple cells. Recent imaging-based spatial transcriptomics platforms enable single-cell resolved expression profiling, revealing how cellular states are organized across different microenvironments and how cells interact with their neighbors~\cite{janesick2023high, chen2015spatially, he2022high}. However, this increased spatial granularity makes H\&E-based prediction more challenging due to sparse transcript counts and gene-specific variability in probe detection efficiency.

Pathology foundation models (FMs) are a promising basis for this task, providing general-purpose representations for histopathology. These models are pretrained at scale on large collections of pathology images, typically using tile-level self-supervised objectives~\cite{dinov2, uni, virchow, provgiga}. Their embeddings capture tissue morphology, but may lack the fine-grained cellular and microenvironmental features needed for single-cell molecular prediction. Because paired H\&E and single-cell spatial transcriptomics data remains scarce, training large-scale models for this task is challenging. Leveraging representations learned by existing pathology FMs and adapting them to capture cell-specific features offers a well-motivated strategy for single-cell level gene expression prediction from H\&E.

In this work, we establish a practical and effective framework for single-cell gene expression prediction using pathology foundation models. Using a large cohort of paired H\&E imaging and Xenium 5K spatial transcriptomics data from patients with inflammatory bowel disease (IBD)---a group of chronic inflammatory disorders associated with an increased risk of colorectal cancer---we systematically implement a series of design choices such as cell-level tokenization, input resolutions, and parameter-efficient fine-tuning to improve single-cell prediction. We show that leveraging pathology FMs as-is works well for spot-level predictions, but is insufficient at single-cell resolution, revealing a key gap from tile-level embeddings to single-cell predictions.

Each of our domain-informed adaptations substantially improves performance across the full gene panel, leading to our model, which integrates all adaptations and achieves state-of-the-art (SOTA) performance in single-cell spatial gene expression prediction. Analyses stratified by biologically informative gene sets, cell types, and pathways reveal substantial heterogeneity in predictability, with particularly strong performance for spatially variable and disease-relevant genes. Together, these results demonstrate that lightweight adaptation enables pathology foundation models to recover biologically meaningful spatial expression patterns at single-cell resolution from H\&E.

\begin{figure}
  \centering
  \includegraphics[width=0.95\linewidth]{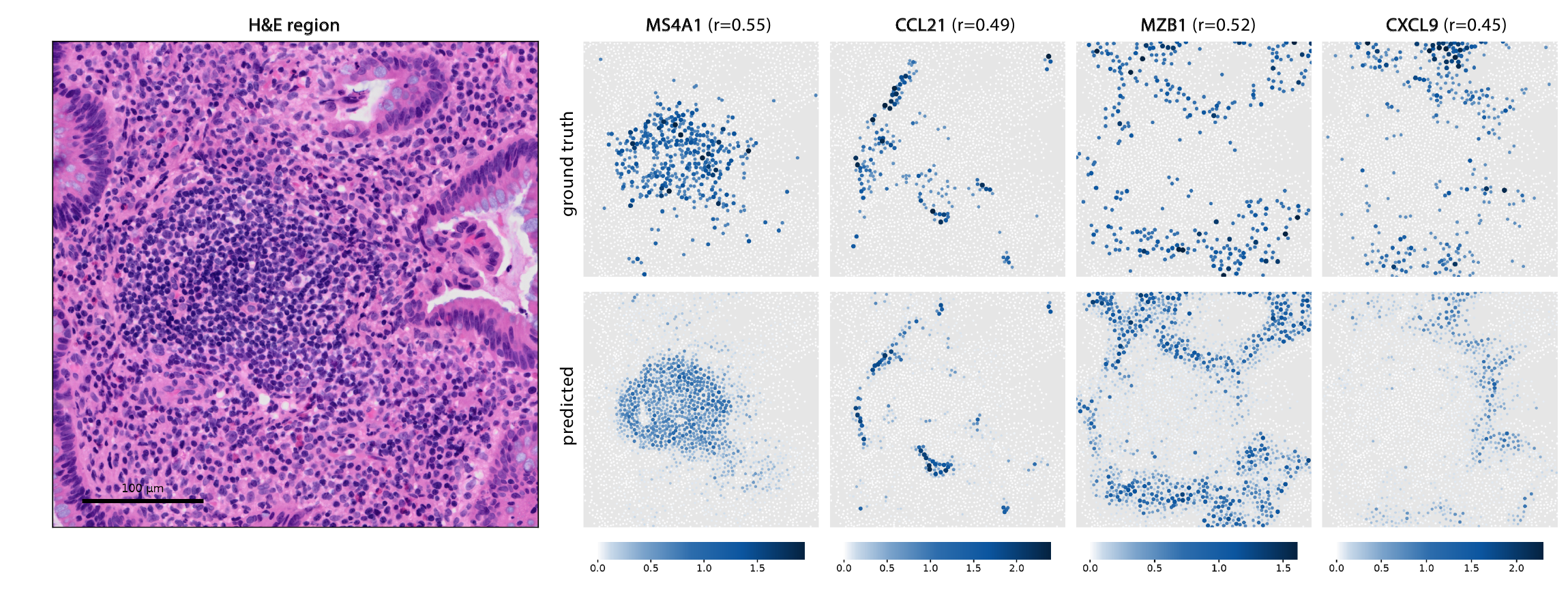}
  \caption{Single-cell predictions from our model in a tertiary lymphoid structure (TLS) region. Our model predicts expression (log1p shown) of the B-cell marker \textit{MS4A1} within the TLS, immune-cell recruitment chemokines \textit{CCL21} and \textit{CXCL9}, and plasma-cell marker \textit{MZB1} at the periphery.}
  \label{fig:tls}
\end{figure}

\section{Related Work}
\label{sec:related}

\paragraph{Pathology foundation models.}
Pathology FMs have emerged as powerful feature extractors for histopathology. Models such as UNI~\cite{uni}, Virchow~\cite{virchow}, Prov-GigaPath~\cite{provgiga}, H-Optimus-0~\cite{hoptimus0} and others~\cite{phikon, atlas2, pluto4} are pretrained at scale on H\&E images and transfer effectively to various downstream tasks such as tissue classification and biomarker prediction. They commonly process fixed-size image tiles and downstream applications often rely on tile-level embeddings~\cite{uni, virchow, atlas2}. These representations capture rich regional morphology but are not explicitly optimized to associate visual features with individual target cells. Our work investigates how the information encoded by these pathology FMs can be localized and adapted for cell-resolved molecular prediction.

\paragraph{Spot-level H\&E to gene expression prediction.}
Prior work established the feasibility of predicting spatial gene expression from histology using paired H\&E and spot-level ST data. ST-Net~\cite{stnet} learns a supervised mapping from H\&E patches to matched expression measurements, while HisToGene~\cite{histogene} and Hist2ST~\cite{hist2st} incorporate transformer-based modeling. Other approaches such as BLEEP~\cite{bleep} and OmiCLIP~\cite{omiclip} explore multi-modal representation learning. HEST-1K~\cite{hest1k} benchmarks zero-shot pathology FMs for gene expression prediction across various ST datasets. Collectively, these studies demonstrate that histology captures expression signals at the regional level, where spot-level aggregation smooths cell-to-cell variation and noise. 

\paragraph{Single-cell H\&E to gene expression prediction.}
Methods such as XFuse~\cite{xfuse} and iStar~\cite{istar} bridge the resolution gap by inferring expression from spot-level ST at super-resolution. sCellST~\cite{scellst} also leverages lower-resolution spot-level ST to predict single-cell level expression. In contrast, our approach and recent methods such as SpatialEx~\cite{spatialex} and GHIST~\cite{ghist} leverage single-cell-resolution ST for supervision. SpatialEx combines hypergraph and contrastive learning, whereas GHIST employs multi-task learning with auxiliary cell-type and neighborhood prediction tasks. Our approach achieves SOTA performance through lightweight adaptations of tile-level pathology foundation models, without requiring specialized architectures or auxiliary biological inputs.
\section{Methods}
\label{sec:methods}

\subsection{Problem formulation}
We formulate single-cell spatial transcriptomics prediction as a supervised learning problem that maps an H\&E tile to its gene expression profile. For each cell \(i \in \{1,\dots,N\}\) with centroid \(c_i\), the input is an RGB H\&E tile $\mathbf{X}_i \in \mathbb{R}^{H \times W \times 3}$ with height $H$ and width $W$ centered at \(c_i\). The target is $\mathbf{y}_i \in \mathbb{R}^{G}$ , where \(G\) is the number of genes to predict. A pretrained pathology foundation model encoder \(f_{\theta_0}\) produces a sequence of token embeddings $\mathbf{H}_i = f_{\theta_0}(\mathbf{X}_i)$. A pooling function $P$ extracts a representation $\mathbf{z}_i = P(\mathbf{H}_i)$, which a prediction head \(g_{\phi}\) maps to a $G$-gene expression vector.

\subsection{Linear probe baseline} 
\label{linear_probe_baseline}
For the baseline, we adapt the HEST-1K~\cite{hest1k} linear probe protocol. We keep \(f_{\theta_0}\) frozen and use its final-layer class token as the representation: $\mathbf{H}_i = f_{\theta_0}(\mathbf{X}_i), P(\mathbf{H}_i) = \mathbf{h}_{i,\mathrm{cls}}$. A linear head predicts log1p-transformed ($\log(1+y)$) gene expression, $\hat{{\mathbf{y}}}_i = \operatorname{Linear}(\mathbf{z}_i)$, and is optimized using mean-squared error (MSE) against the log1p-transformed target. We extract an H\&E patch centered on the cell centroid, spanning a physical field of view (FOV) of 112$\times$112$~\mu\mathrm{m}$ at 20\(\times\) magnification. This resolution is commonly used in pretrained pathology FMs and downstream tasks~\cite{ctranspath, phikon, uni, stnet}. The resulting embedding is used to predict the cell's corresponding gene expression profile.

\subsection{Nonlinear prediction head}
While the linear probe baseline maps the image representation to gene expression using a single linear layer, we replace this with a nonlinear prediction head. $\mathbf{z}_i$ is passed to a two-layer multilayer perceptron (MLP) prediction head, $g_{\phi}$, with a 512-dimensional hidden layer and ReLU activation.

\subsection{Core design choices for single-cell adaptation} 
\label{single_cell_adaptation}

\begin{figure}
  \centering
  \includegraphics[width=0.98\linewidth]{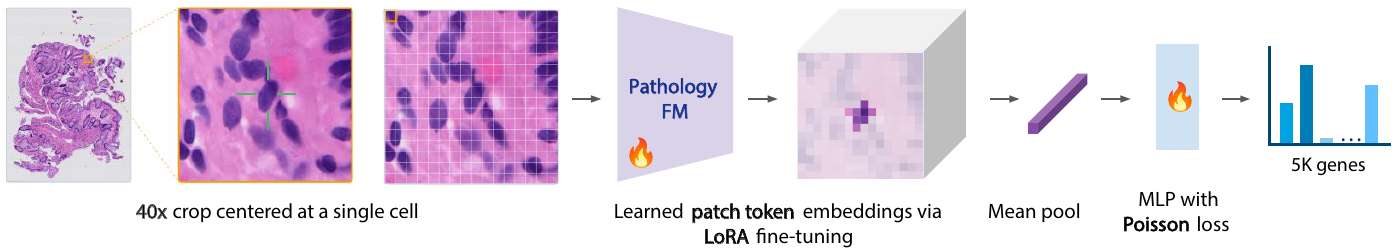}
  \caption{Framework for predicting single-cell gene expression from H\&E images. We integrate domain-informed design choices including high-resolution inputs, parameter-efficient encoder fine-tuning, cell-localized features, and a Poisson-trained MLP to predict expression across 5K genes.}
  \label{fig:workflow}
\end{figure}

Building on the nonlinear prediction head, we introduce a series of lightweight adaptations tailored to single-cell spatial transcriptomics and gene-expression prediction from H\&E images (Figure~\ref{fig:workflow}).

\paragraph{Cell-based feature pooling.}
Many pathology foundation models use the class token as a global representation of an H\&E tile~\cite{uni,virchow,atlas2}. For single-cell prediction, however, we use the granular patch-token embeddings associated with the cell of interest. Let $\mathbf{H}_i = \{\mathbf{h}_{i,1},\ldots,\mathbf{h}_{i,T}\}$ denote the patch-token embeddings for image patch $\mathbf{X}_i$, and let $u_{i,t}$ denote the spatial center of token $t$. Given the Xenium-derived cell mask $M_i$, we define the token-membership indicator $m_{i,t} = \mathbf{1}\{u_{i,t} \in M_i\}$, which selects tokens whose spatial centers fall inside the mask. For cells with no token centers inside $M_i$, we instead select the single token whose center is nearest to $c_i$, so that at least one token is always selected. We then obtain the cell representation by mean pooling the selected token embeddings: $P(\mathbf{H}_i) = \frac{\sum_{t=1}^{T} m_{i,t}\mathbf{h}_{i,t}}{\sum_{t=1}^{T} m_{i,t}}$.

\paragraph{Input resolution.}
Spot-level prediction typically uses 224$\times$224-pixel H\&E tiles at 0.5$~\mu\mathrm{m}$/pixel (20$\times$), corresponding to a 112$\times$112$~\mu\mathrm{m}$ FOV~\cite{stnet, ctranspath, uni, phikon}. For single-cell prediction, we instead extract patches spanning a 47.6$\times$47.6$~\mu\mathrm{m}$ FOV at 40$\times$ resolution. This configuration provides finer spatial detail around the target cell while retaining local tissue context.

\paragraph{Loss function.}
Because Xenium expression measurements are sparse, non-negative counts, we train on raw transcript counts using the Poisson negative log-likelihood rather than the MSE on log1p-transformed targets~\cite{zhao2022modeling, sarkar2021separating}. The model predicts gene-specific rates $\boldsymbol{\lambda}_i = \mathrm{softplus}(g_{\phi}(\mathbf{z}_i))$, where the softplus activation ensures that all predicted rates are positive. The training objective is 
\[
  \mathcal{L}_{\mathrm{Poisson}}
  =
  \frac{1}{NG}
  \sum_{i=1}^{N}
  \sum_{g=1}^{G}
  \left(
      \lambda_{ig}
      -
      y_{ig}\log \lambda_{ig}
      +
      \log\!\left(y_{ig}!\right)
  \right),
\]
where $y_{ig}$ is the raw transcript count for gene $g$ in cell $i$. For evaluation, predicted rates and observed counts are log1p-transformed.

\paragraph{Parameter-efficient fine-tuning.}
Since pathology FMs were not directly trained for cell-resolved transcriptomic prediction, we further augment \(f_{\theta_0}\) with Low-Rank Adaptation (LoRA)~\cite{lora} parameters $\psi$, producing the histology representation $\mathbf{H}_i = f_{\theta_0, \psi}(\mathbf{X}_i)$. Keeping $\theta_0$ fixed, we optimize only the LoRA adapters $\psi$ and the prediction head. We add LoRA adapters to the query, key, and value projections in every self-attention block, using rank $r$ = 16, scaling factor $\alpha$ = 32, and adapter dropout 0.05. The loss is computed only from the cell-localized tokens; since self-attention operates globally over the image tile, these tokens nevertheless incorporate information from the entire tile.

\section{Experiments}
\label{sec:experiments}

\subsection{Experimental setup}

\paragraph{Dataset.}
\label{sec:data}
We used two colon tissue datasets with paired H\&E imaging and 10x Genomics Xenium 5K Human Colon profiling. H\&E images were acquired at 40$\times$ resolution (0.2125$~\mu\mathrm{m}$/pixel), registered to the Xenium data, and paired with cell boundaries generated by the standard on-instrument segmentation pipeline. Of the 5,399 profiled genes, zero-variance genes were excluded, leaving 5,101 genes for training and evaluation. 

The tissue microarray (TMA) cohort comprised 176 ulcerative colitis (UC) cores, spanning 117 donors, 10 TMA blocks, and totaling about 8.6 million cells. All cores from a given donor were placed on the same TMA block; thus, a block-level splitting strategy eliminates biological leakage from donor-associated effects and minimizes technical leakage from block-associated effects. Five blocks (85 cores) were used for fine-tuning the image encoder, and five (91 cores) for downstream evaluation. Evaluation used five-fold cross-validation at the block level: three blocks for training, one for validation, and one for testing. The resection cohort comprised two UC, two Crohn's disease, and two healthy control samples, totaling about 2.9 million cells, and were used for fine-tuning only.

\paragraph{Metrics.}
Performance on held-out cells was evaluated primarily using gene-wise Pearson correlation coefficient (PCC), computed between log1p-transformed predicted and measured expression across cells~\cite{bleep, scellst, wang2025benchmarking}. For each gene panel, we report mean PCC across its genes. We also compute per-cell PCC across all genes and report the mean across cells. Gene-detection performance was assessed using gene-wise area under the receiver operating characteristic curve (AUROC), with measured counts binarized as detected $(y_{ig} > 0)$, and corresponding continuous predictions used as scores. Metrics were computed separately for each test fold and summarized as mean \(\pm\) standard deviation. Full metric definitions are provided in Appendix~\ref{app:metrics}.

\paragraph{Evaluations.}
We evaluate four gene panels for comprehensive evaluation: all 5,101 genes (5K), the top 50 highly variable genes (HVG50), the top 50 spatially variable genes (SVG50), and 76 literature-derived IBD-relevant genes (IBD)~\cite{garridotrigo2023macrophage, maciag2024jakstat, mcgregor2025spatial, elmentaite2021cells, theocharidis2022single}. HVG50 and SVG50 were selected independently from each fold's training data. HVGs were ranked by normalized dispersion using Scanpy~\cite{scanpy}, and SVGs by Moran's I~\cite{morans}, both using log1p-transformed raw counts. 

\paragraph{Implementation details.}
The following experiments use H-Optimus-0~\cite{hoptimus0}, a pathology foundation model based on a ViT-g/14 ~\cite{vit} with a DINOv2~\cite{dinov2} backbone, which generates 1536-dimensional feature representations. Prediction heads were trained independently within each fold using AdamW~\cite{adamw}. Early stopping and model selection used mean per-gene PCC on the validation split, and the best checkpoint was evaluated once on the held-out test split. LoRA~\cite{lora} adaptation used a separate cohort  disjoint from all evaluation folds (Section \ref{sec:data}). As with $f_{\theta_0}$, we use $f_{\theta_0, \psi}$ as a frozen feature extractor and train the prediction head using the same cross-validation protocol. Full optimization and LoRA settings are provided in Appendix~\ref{app:implementation}.

\begin{table}[t]
\centering
\caption{Mean per-gene PCC across cumulative model adaptations. Values in parentheses indicate relative improvement over the preceding configuration. $^{\dagger}$Configuration: linear probe, class-token embeddings, 20$\times$ input, MSE loss, and frozen embeddings.}
\label{tab:config-ladder}
\small
\setlength{\tabcolsep}{8pt}

\begin{tabular}{lcccc}
\toprule
Model & SVG50 & IBD & HVG50 & 5K \\
\midrule

OmiCLIP~\cite{omiclip}
& 0.207
& 0.072
& 0.057
& 0.029 \\

BLEEP~\cite{bleep}
& 0.349
& 0.160
& 0.117
& 0.065 \\

GHIST~\cite{ghist}
& 0.505
& 0.167
& 0.087
& 0.081 \\

SpatialEx~\cite{spatialex}
& 0.509
& 0.207
& 0.158
& 0.094 \\

\midrule

Baseline$^{\dagger}$
& 0.469
& 0.182
& 0.151
& 0.065 \\

+ MLP
& 0.525 {\scriptsize($+12.0\%$)}
& 0.235 {\scriptsize($+28.8\%$)}
& 0.195 {\scriptsize($+29.5\%$)}
& 0.110 {\scriptsize($+69.0\%$)} \\

+ cell\_pool
& 0.583 {\scriptsize($+10.9\%$)}
& 0.285 {\scriptsize($+21.6\%$)}
& 0.253 {\scriptsize($+29.5\%$)}
& 0.135 {\scriptsize($+22.6\%$)} \\

+ 40$\times$
& 0.602 {\scriptsize($+3.3\%$)}
& 0.298 {\scriptsize($+4.4\%$)}
& 0.248 {\scriptsize($-1.8\%$)}
& 0.146 {\scriptsize($+7.8\%$)} \\

+ Poisson
& 0.618 {\scriptsize($+2.7\%$)}
& 0.324 {\scriptsize($+8.8\%$)}
& 0.301 {\scriptsize($+21.0\%$)}
& 0.161 {\scriptsize($+10.6\%$)} \\

+ LoRA (Ours)
& \textbf{0.643} {\scriptsize($\mathbf{+4.0\%}$)}
& \textbf{0.356} {\scriptsize($\mathbf{+9.7\%}$)}
& \textbf{0.338} {\scriptsize($\mathbf{+12.5\%}$)}
& \textbf{0.174} {\scriptsize($\mathbf{+7.9\%}$)} \\

\addlinespace
Baseline $\rightarrow$ Ours
& $\mathbf{+37.1\%}$
& $\mathbf{+95.6\%}$
& $\mathbf{+124.2\%}$
& $\mathbf{+166.7\%}$ \\

\bottomrule
\end{tabular}
\end{table}

\subsection{Pathology foundation models do not natively transfer to single-cell expression prediction}

We first assessed how well frozen pathology foundation models transfer to gene expression prediction at spot and single-cell resolution. Following the HEST-1K linear-probing framework~\cite{hest1k}, we evaluated both tasks using the same Xenium data. To construct Visium-like pseudo-spots, we assigned cells to a regular grid of 55$\times$55$~\mu\mathrm{m}$ bins, approximating the diameter of a Visium spot. For each pseudo-spot, raw gene counts were summed and log1p-transformed. Each pseudo-spot was paired with a centered 112$\times$112$~\mu\mathrm{m}$ H\&E tile, resized to 224$\times$224 pixels at 0.5$~\mu\mathrm{m}$/pixel, to predict its corresponding pseudo-bulked expression profile. On HVG50, the mean per-gene PCC was 0.428, comparable to the spot-level performance reported by HEST-1K~\cite{hest1k}.

\begin{figure}
  \centering
  \includegraphics[width=0.95\linewidth]{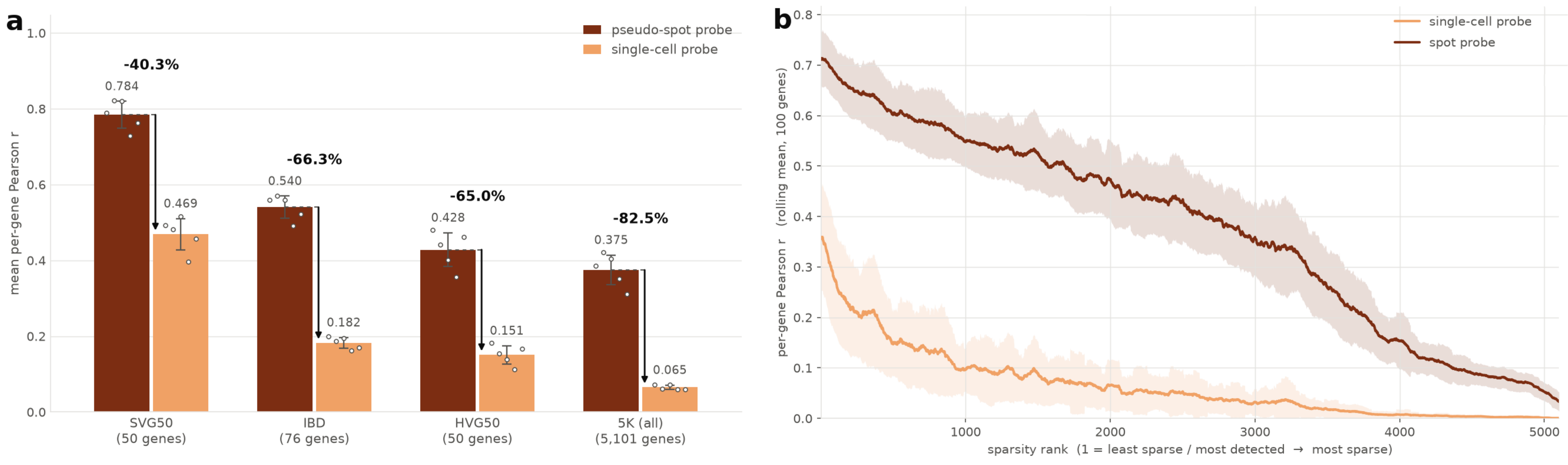}
  \caption{Baseline performance with frozen pathology foundation model embeddings at pseudo-spot and single-cell resolutions: (a) across gene panels; (b) per-gene over increasing expression sparsity.}
  \label{fig:spot_vs_sc}
\end{figure}

Linear probes on frozen pathology FM embeddings performed substantially better at the pseudo-spot level than at the single-cell level, suggesting that these embeddings capture histology-associated gene expression patterns more effectively at regional than at cellular resolution. For the full 5K panel, mean per-gene PCC decreased from 0.375 at pseudo-spot resolution to 0.065 at single-cell resolution. Similar declines were observed for the other gene panels (Figure \ref{fig:spot_vs_sc}a), although performance varied substantially across panels, with SVG50 achieving the highest PCCs. Together, these results suggest that native pathology FM embeddings are better suited to regional than cell-resolved expression prediction, with performance depending on both spatial resolution and gene-panel composition. We further observed that PCC declined with increasing gene expression sparsity. At the spot-level, performance decreased gradually, whereas at the single-cell level, the decline was substantially more pronounced (Figure \ref{fig:spot_vs_sc}b). This pattern highlights the greater challenge of inferring sparse gene expression at single-cell resolution using native pathology foundation model embeddings.

Given these limitations, we replaced the linear probe with a two-layer MLP decoder, which improved performance across all gene panels (Table \ref{tab:config-ladder}, +MLP). These gains suggest that frozen FM embeddings retain expression-relevant information that is better recovered by a higher-capacity nonlinear decoder, and standard linear probes may be insufficiently expressive for single-cell expression prediction.

\subsection{Single-cell-specific adaptations yield substantial cumulative performance gains}

\paragraph{Cell-localized token pooling improves single-cell representations.}
Pooling tokens within cell boundaries substantially improved performance relative to the class token baseline (Table \ref{tab:config-ladder}, +cell\_pool). We also evaluated centroid-based pooling strategies that do not require cell-boundary masks. For each cell, we selected tokens within a radius of 5$\,\mu \mathrm{m}$ from its centroid and aggregated them using mean pooling or distance-weighted pooling (Appendix~\ref{app:cell-token-pooling}, Table~\ref{tab:cell_token}). Compared with mask-based pooling, distance-weighted and mean pooling reduced PCC for 5K by only 4.6\% and 7.3\%, respectively. Both approaches nevertheless substantially outperformed the class token baseline, where distance-weighted and mean pooling improved PCC by 20.2\% and 16.4\%, respectively. 

These results show that restricting representations to the target cell and its immediate surroundings is more effective than using a global tile-level embedding, and that much of this benefit can be retained without explicit cell boundaries. In an H\&E-only deployment setting, cell nuclei or boundaries would need to be inferred using a segmentation model~\cite{cellvit, cellvit++, cellsam}. To simulate this setting without the Xenium-derived cell information, we used CellViT++~\cite{cellvit++} to detect nuclei centroids in H\&E images and found high concordance with the Xenium-derived nuclei coordinates (Appendix~\ref{app:he-derived-centroids}). Together with the strong performance of distance-weighted pooling, these findings support the feasibility of nuclei-based cell-localized pooling in H\&E-only settings, while highlighting the potential for further gains from explicit cell boundaries.
 
\paragraph{40\(\times\) inputs balance cellular detail with local tissue context.}

40\(\times\) inputs (47.6$\times$47.6$~\mu\mathrm{m}$ FOV) improved performance compared to 20$\times$ inputs (Table \ref{tab:config-ladder}, +40$\times$). At this resolution, each cell spans more image pixels and spatial tokens, providing a richer representation of cellular and subcellular morphology, while also retaining local tissue context around the target cell (Appendix~\ref{app:input-res}, Figure~\ref{fig:fovs}). In comparison, using a broader FOV at 10$\times$ reduced PCC on 5K by 19.8\% relative to 40$\times$ (Appendix~\ref{app:input-res}, Table \ref{tab:input-resolution}), likely because the surrounding tissue dilutes target-cell-specific signals. On the other extreme, isolated single-cell crops reduced PCC by 15.5\% (Appendix~\ref{app:input-res}, Table \ref{tab:input-resolution}). Such crops remove potentially informative microenvironmental context and may also be poorly matched to pathology FM pretraining, which typically captures tiles of tissue regions rather than isolated cells. 

\paragraph{Poisson count modeling improves single-cell expression prediction.}
Poisson supervision improved performance across all gene subsets (Table \ref{tab:config-ladder}, +Poisson), with the largest gain observed for HVG50. These results indicate that modeling the raw transcript counts directly provides a more informative supervisory signal for single-cell prediction than regression on transformed expression values. Compared with a Gaussian regression, the Poisson objective better reflects the discrete nature and mean–variance relationship of count data, consistent with previous findings~\cite{zhao2022modeling, sarkar2021separating}.

\paragraph{Fine-tuning further aligns H\&E features with single-cell expression.}
LoRA fine-tuning improved mean per-gene PCC across all gene subsets (Table \ref{tab:config-ladder}, +LoRA). While our frozen adapted model already translated well to the task, parameter-efficient task adaptation further aligned the features with single-cell gene expression prediction (Appendix~\ref{app:umap}, Figure~\ref{fig:umap}). 

Our adapted model substantially outperformed the pre-adaptation baseline, improving performance by 166.7\% on 5K, 124.2\% on HVG50, 95.6\% on IBD, and 37.1\% on SVG50, and also achieved SOTA performance for this task (Table ~\ref{tab:config-ladder}). The single-cell predictions from our model accurately recapitulated expression patterns within the tertiary lymphoid structure in colon tissue, including the spatial localization of relevant immune and signaling markers (Figure~\ref{fig:tls}). Our model also demonstrated significant strengths in predicting epithelial marker genes, across both healthy and diseased samples (Appendix~\ref{app:epi-preds}, Figure~\ref{fig:epi_preds}).

\subsection{Our adaptations generalize across foundation models}
\begin{figure}
  \centering
  \includegraphics[width=0.95\linewidth]{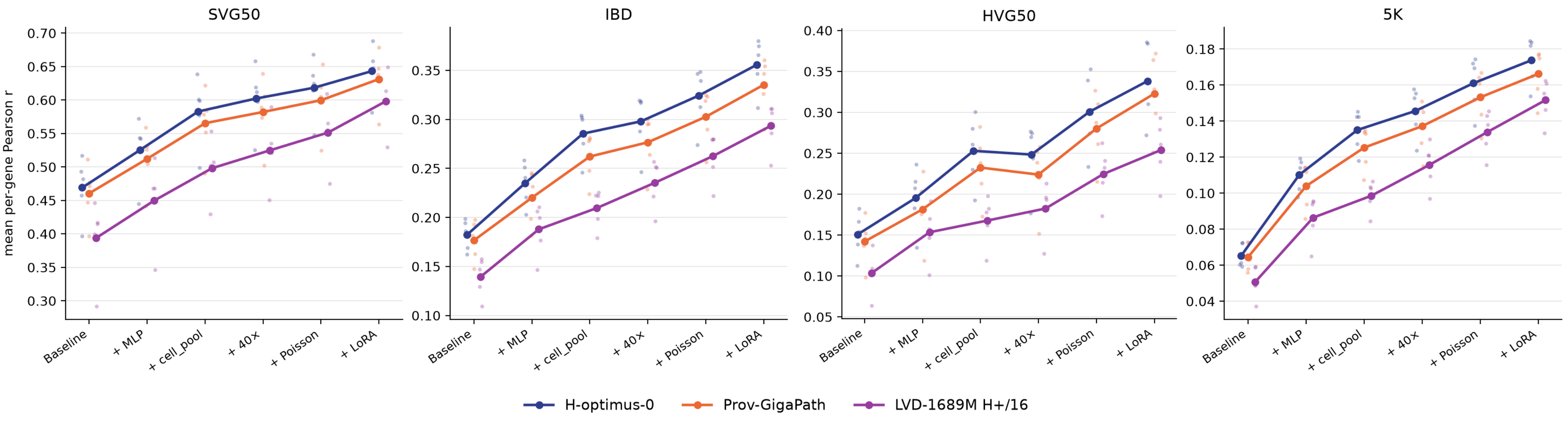}
  \caption{Our domain-driven model adaptations generalize across different foundation models.}
  \label{fig:rung_comp}
\end{figure}

We applied these domain-driven adaptations to two additional foundation models to evaluate if these improvements generalize beyond H-Optimus-0~\cite{hoptimus0}. Prov-GigaPath~\cite{provgiga} was selected as another pathology FM due to its comparable use of DINOv2, a 14$\times$14 patch size, and an embedding dimension of 1536. We also evaluated the distilled DINOv3~\cite{simeoni2025dinov3} model trained on the non-histopathology LVD-1689M dataset, which uses a 16$\times$16 patch size and a 1280 embedding dimension. We observe that our improvements successfully transfer regardless of the foundation model used (Figure~\ref{fig:rung_comp}).

\subsection{Our adapted single-cell model outperforms existing H\&E-to-ST methods}

We benchmarked our model against several existing approaches for spatial transcriptomics prediction from H\&E, including SpatialEx~\cite{spatialex}, GHIST~\cite{ghist}, OmiCLIP~\cite{omiclip}, and BLEEP~\cite{bleep}. SpatialEx and GHIST predict single-cell gene expression using supervision from single-cell-resolution ST data. OmiCLIP is a visual--omics foundation model trained on Visium-scale ST and BLEEP introduced a contrastive retrieval framework for spot-level prediction from H\&E. For BLEEP, we lowered the contrastive temperature to $\tau$ = 0.07 to obtain a stable baseline for single-cell predictions. OmiCLIP was evaluated in a zero-shot setting, while GHIST and SpatialEx were  trained end-to-end. For GHIST, its cell-type, neighborhood, and reference-expression components were disabled for a fair comparison with the intended H\&E-only constraint. 

Our adapted model, which combines all the task-specific adaptations, achieved the best performance across all evaluated metrics (Table~\ref{tab:benchmark-5k}). It outperformed SpatialEx, the strongest competing method, on all three metrics on the 5K panel: mean per-gene PCC (0.174$\pm$0.014 vs. 0.094$\pm$0.009), mean per-cell PCC (0.280$\pm$0.033 vs. 0.242$\pm$0.031), and AUROC (0.800$\pm$0.009 vs. 0.642$\pm$0.005). Even without LoRA fine-tuning, our frozen adapted model ranked second across all metrics, demonstrating that the single-cell architectural adaptations alone substantially improve upon existing approaches. Although precise spatial expression prediction across 5K genes at single-cell resolution remains challenging, our method consistently outperformed competing approaches, with its strong gene-detection AUROC also indicating reliable detection of expression signals.

\begin{table}
  \centering
  \caption{Benchmarking single-cell gene expression predictions from H\&E
  on the 5K panel. We report mean per-gene PCC, mean per-cell PCC, and
  AUROC. Native granularity indicates the resolution-level the model was trained on; spot-level methods were evaluated on the single-cell prediction task.}
  \label{tab:benchmark-5k}
  \small
  \setlength{\tabcolsep}{7pt}
  \begin{tabular}{llccc}
  \toprule
  Model & Native granularity & PCC (per-gene) & PCC (per-cell) & AUROC \\
  \midrule
  OmiCLIP~\cite{omiclip}
  & Spot-level
  & 0.029 $\pm$ 0.006
  & 0.203 $\pm$ 0.026
  & 0.568 $\pm$ 0.016 \\

  BLEEP~\cite{bleep}
  & Spot-level
  & 0.065 $\pm$ 0.041
  & 0.124 $\pm$ 0.078
  & 0.578 $\pm$ 0.062 \\

  GHIST~\cite{ghist}
  & Single-cell
  & 0.081 $\pm$ 0.007
  & 0.242 $\pm$ 0.027
  & 0.605 $\pm$ 0.012 \\
  SpatialEx~\cite{spatialex}
  & Single-cell
  & 0.094 $\pm$ 0.009
  & 0.242 $\pm$ 0.031
  & 0.642 $\pm$ 0.005 \\

  Ours (frozen)
  & Single-cell
  & 0.161 $\pm$ 0.016
  & 0.271 $\pm$ 0.034
  & 0.786 $\pm$ 0.005 \\

  Ours (LoRA)
  & Single-cell
  & \textbf{0.174 $\pm$ 0.014}
  & \textbf{0.280 $\pm$ 0.033}
  & \textbf{0.800 $\pm$ 0.009} \\
  \bottomrule
  \end{tabular}
\end{table}

\subsection{Aggregated single-cell predictions improve spot-level expression prediction}

At single-cell resolution, our model achieved higher per-gene PCCs across every gene in the 5K panel and consistently outperformed the baseline across levels of gene expression sparsity (Appendix~\ref{app:aggregated-preds}, Figure \ref{fig:spot_vs_sc_combined}a). To assess whether cell-level predictions recovered coherent regional expression patterns, we aggregated predicted and measured single-cell expression into 55\(~\mu\mathrm{m}\) and 100\(~\mu\mathrm{m}\) pseudo-spots. Values were converted to counts, summed within each spot, and log1p-transformed. Aggregation improved correlations overall, and our model maintained a substantial advantage for every gene at both resolutions (Appendix~\ref{app:aggregated-preds}, Figure \ref{fig:spot_vs_sc_combined}b, c). These results demonstrate that our model, optimized for single-cell prediction, also yields more accurate spatially aggregated expression profiles.

\subsection{Cell-type-stratified analysis reveals populations that benefit most from our approach}

We next investigated whether our adaptations preferentially benefited specific cell types. We computed the per-cell PCC across all genes and averaged these scores within each annotated cell type. Cell-type labels were obtained using CONCORD~\cite{concord}, which jointly embeds a reference scRNA-seq atlas with the ST data, followed by annotation with a calibrated k-nearest-neighbor classifier (Appendix~\ref{app:cell-type-coverage}).

\begin{figure}
  \centering
  \includegraphics[width=0.95\linewidth]{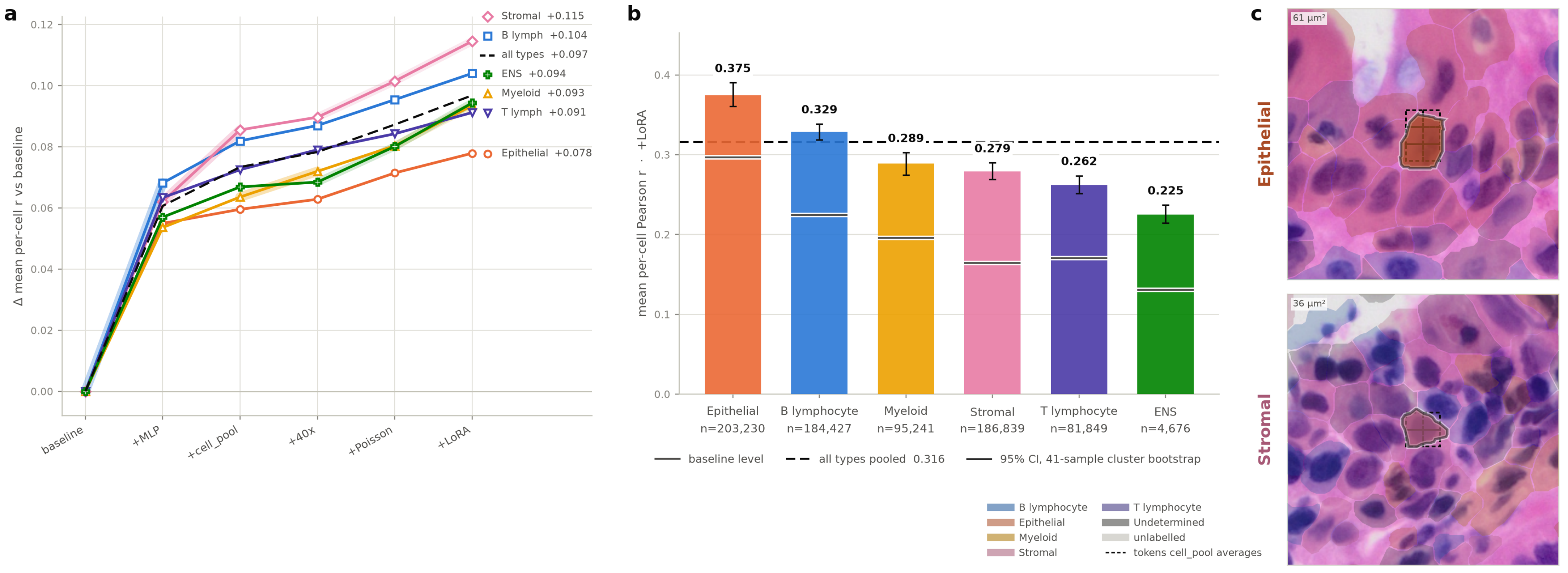}
  \caption{Cell-type-specific effects of single-cell modeling adaptations. (a) Gains in performance across adaptations. (b) Performance stratified by cell type. (c) Example H\&E image regions.}
  \label{fig:ct_analysis}
\end{figure}

Cell-localized pooling particularly benefited stromal cells (Figure \ref{fig:ct_analysis}a). Overall, stromal cells showed the largest gains from baseline to our model, while epithelial cells showed the smallest. Epithelial cells already performed relatively well at baseline (Figure \ref{fig:ct_analysis}b), possibly because their cohesive organization and distinct architecture provide strong morphological cues in H\&E images~\cite{villanacci2021histopathology}. In contrast, stromal cells are heterogeneous and interact with diverse cellular and extracellular components in their microenvironment~\cite{barnhoorn2020stromal} (Figure \ref{fig:ct_analysis}c), potentially making them more sensitive to the inclusion of features specific to the target cell. The global class token may capture signals from the surrounding mixture of cells, whereas cell-localized pooling emphasizes target-cell features. For epithelial cells, these signals may be more concordant, limiting the additional benefit of localized pooling. T lymphocytes also showed relatively small gains, potentially because their small size and relatively uniform nuclear morphology in H\&E provide limited information for molecular discrimination~\cite{cano2013introduction}.

These results suggest that single-cell adaptations, such as cell-token pooling, are beneficial for cell types whose expression-associated features are poorly isolated by global, tile-level representations. More broadly, they emphasize the importance of evaluating performance across biologically informed populations rather than relying exclusively on aggregate metrics, which can obscure cell-type-specific improvements and persistent performance gaps.
    
\subsection{Pathway analysis identifies biological programs associated with single-cell adaptations}

To identify the biological programs that benefited most from single-cell adaptations, we ranked genes by the change in mean per-gene PCC relative to the baseline. We performed REACTOME~\cite{reactome} pathway enrichment analysis on the 500 most improved and 500 least improved genes using g:Profiler~\cite{gprofiler} (details in Appendix~\ref{app:pathway}).

The most improved genes from our model were enriched for pathways related to extracellular matrix organization, cell junctions, adhesion, and cell-cell communication pathways (Appendix~\ref{app:pathway}, Figure \ref{fig:pathways}a). These processes are closely linked to tissue architecture and local cell--cell interactions, suggesting that the adaptations particularly improve prediction of genes with visible morphological or microenvironmental correlates. This is also consistent with our cell-type-stratified analysis, in which stromal populations exhibited large performance gains following single-cell adaptations.

The least improved genes were enriched for G-protein-coupled receptor (GPCR) signaling and neurotransmitter-related pathways (Appendix~\ref{app:pathway}, Figure \ref{fig:pathways}b). These may represent signaling states that are transient, weakly reflected in H\&E morphology, or dependent on molecular context that cannot be inferred from histology alone. Enrichment among the least-improved genes may also be influenced by technical and statistical factors, including sparsity, limited dynamic range, or low detection rates. We therefore interpret these enrichment results as hypothesis-generating rather than as definitive evidence that particular biological programs are or are not morphologically predictable.

\section{Conclusion}
\label{sec:conclusion}

In this work, we demonstrate that a set of domain-driven, task-specific design choices achieves state-of-the-art performance for predicting single-cell gene expression from H\&E images. We found that frozen pathology foundation model embedding approaches transfer poorly to cell-resolved targets, motivating adaptations that better reflect the spatial resolution, sparsity, and complexity of single-cell data. Nonlinear decoding with additional parameters, cell-localized spatial-token pooling, 40$\times$ inputs that preserve both cellular detail and local tissue context, Poisson supervision, and LoRA fine-tuning together establish substantial improvements over existing methods for this task.

We further showed that aggregate performance metrics alone are insufficient to characterize this task. By evaluating performance across gene panels, expression-sparsity levels, cell types, spatial aggregation scales, and biological pathways, we identified where single-cell adaptation provides the greatest benefit and where important performance gaps remain. The modest overall correlation partly reflects the sparsity and  detection variability of single-cell transcript measurements, which can limit observed correlations despite meaningful biological signals. Predictive performance varies substantially across genes, with notably higher Pearson correlations for spatially variable genes ($r$ = 0.643) and IBD-associated genes ($r$ = 0.356), compared with the full 5K panel ($r$ = 0.174). Our model recovers spatial expression patterns of key immune and epithelial markers, suggesting its potential to characterize tissue microenvironments and spatially organized molecular programs.

Our work has several limitations. Our experiments relied on single-cell imaging-based ST technologies, which are susceptible to transcript spillover, imperfect cell segmentation, and cell-assignment errors. Further evaluation is needed to determine whether these predictions capture subtle cell states and within-cell-type molecular variation beyond broad cell identity. In addition, broader validation across tissues and cohorts is needed to establish generalizability. Future work should further characterize model uncertainty, particularly on a per-gene and per-cell-type basis. Although the relatively low correlation scores in the 5K predictions indicate the difficulty of predicting single-cell gene expression from H\&E, our downstream cell type and pathway enrichment analyses demonstrate that the resulting predictions can provide meaningful insights into the model's behavior and capabilities. 

Such analyses will be essential for determining when molecular profiles inferred from H\&E are sufficiently accurate, reliable, and useful in research and translational settings. Overall, our work establishes a foundation for adapting pathology foundation models to single-cell spatial transcriptomics. We demonstrate that carefully chosen, lightweight modifications to baseline foundation model approaches---rather than extensive architectural or methodological redesign---can enable scalable and interpretable molecular profiling from routine histology.

\begin{ack}

We are grateful to Alma Andersson for feedback on the manuscript and valuable input, and Rushin Gindra for insightful discussions.
\end{ack}

\bibliographystyle{unsrt}
\bibliography{egbib}


\newpage
\appendix
\section{Appendix}
\label{appendix}

\subsection{Formal metric definitions}
\label{app:metrics}
For each gene \(g\), we compute the Pearson correlation (PCC) across cells in the evaluation set:
\[
    r_g =
    \frac{
        \sum_{i=1}^{N_{\mathrm{test}}}
        \left(\hat{y}_{ig} - \bar{\hat{y}}_{g}\right)
        \left(y_{ig} - \bar{y}_{g}\right)
    }{
        \sqrt{
            \sum_{i=1}^{N_{\mathrm{test}}}
            \left(\hat{y}_{ig} - \bar{\hat{y}}_{g}\right)^2
        }
        \sqrt{
            \sum_{i=1}^{N_{\mathrm{test}}}
            \left(y_{ig} - \bar{y}_{g}\right)^2
        }
    },
\]
where \(y_{ig}\) and \(\hat{y}_{ig}\) denote the measured and predicted expression values for gene \(g\) in cell \(i\), respectively, and \(\bar{y}_{g}\) and \(\bar{\hat{y}}_{g}\) denote the corresponding means across the evaluation cells. Overall performance for a gene set \(\mathcal{G}\) is summarized as the mean PCC across genes,
\[
    \mathrm{PCC}(\mathcal{G})
    =
    \frac{1}{|\mathcal{G}|}
    \sum_{g \in \mathcal{G}} r_g.
\]

To inform correlation in a cell-centric manner, we also compute PCC at the cell level. For cell \(i\), the per-cell PCC is
\[
    r_i =
    \frac{
        \sum_{g \in \mathcal{G}}
        \left(\hat{s}_{ig} - \bar{\hat{s}}_{i}\right)
        \left(s_{ig} - \bar{s}_{i}\right)
    }{
        \sqrt{
            \sum_{g \in \mathcal{G}}
            \left(\hat{s}_{ig} - \bar{\hat{s}}_{i}\right)^2
        }
        \sqrt{
            \sum_{g \in \mathcal{G}}
            \left(s_{ig} - \bar{s}_{i}\right)^2
        }
    },
\]
where \(\bar{s}_{i}\) and \(\bar{\hat{s}}_{i}\) denote the mean measured and
predicted expression across genes in \(\mathcal{G}\) for cell \(i\). 

To assess whether models correctly identify gene detection events, we compute AUROC for each gene by binarizing measured expression as
\[
    a_{ig} = \mathbb{I}(y_{ig} > 0),
\]
where \(a_{ig}=1\) indicates that gene \(g\) was detected in cell \(i\). The continuous prediction score was the predicted expression value
\(\hat{s}_{ig}\). For each gene \(g\), let
\(\mathcal{P}_g = \{i : a_{ig}=1\}\) and
\(\mathcal{N}_g = \{i : a_{ig}=0\}\). The gene-wise AUROC is
\[
    \mathrm{AUROC}_g
    =
    \frac{1}{|\mathcal{P}_g| |\mathcal{N}_g|}
    \sum_{i \in \mathcal{P}_g}
    \sum_{j \in \mathcal{N}_g}
    \left[
        \mathbb{I}(\hat{s}_{ig} > \hat{s}_{jg})
        +
        \frac{1}{2}\mathbb{I}(\hat{s}_{ig} = \hat{s}_{jg})
    \right].
\]

AUROC for a gene set \(\mathcal{G}\) is summarized as
\[
    \mathrm{AUROC}_{\mathcal{G}}
    =
    \frac{1}{|\mathcal{G}|}
    \sum_{g \in \mathcal{G}}
    \mathrm{AUROC}_g.
\]

Genes with only one detected class in an evaluation fold, i.e. all detected or
all undetected, were excluded from AUROC calculation for that fold.

\subsection{Implementation full details}
\label{app:implementation}

Heads are optimized with AdamW~\cite{adamw} at a learning rate of \(10^{-3}\), batch size 8192, for at most 30 epochs with gradient clipping at 1.0 and mixed precision, on a single A100 GPU. Model selection and early stopping use mean per-gene PCC on the fold's validation split (patience 5 epochs); the best-scoring checkpoint is restored before the held-out test split is scored once.

LoRA~\cite{lora} is run on the fine-tuning cohort, disjoint from the evaluation folds. The encoder and the head use separate learning rates (\(10^{-4}\) and \(10^{-3}\)), with 500 steps of linear warm-up followed by cosine decay to \(10^{-6}\), AdamW, batch size 128 with gradient accumulation over 2 steps (effective batch 256), gradient clipping at 1.0, and a cap of 30{,}000 optimizer steps. We early-stop on validation mean per-gene PCC with patience 4 and a minimum improvement of 0.001. The adapted encoder is then frozen and used as a feature extractor: all cells are re-embedded, and the readout is refit under exactly the same \(k\)-fold protocol as the frozen encoders, so the adaptation and frozen rungs differ only in the encoder weights.

\subsection{Cell-token pooling experiments}
\label{app:cell-token-pooling}
Tokens within a fixed 5$\mu\mathrm{m}$ radius of the centroid (12 of 256 tokens at 40$\times$) were pooled either uniformly or using Gaussian distance weights, \(w_k = \exp(-\tfrac{1}{2}(d_k/\sigma)^2)\), with $\sigma$ = 2.5$\mu\mathrm{m}\,$. Distance-weighted pooling performed slightly worse than mask-guided pooling (PCC = 0.154 vs. 0.161) but better than uniform radius pooling (PCC = 0.149) and the class-token baseline (PCC = 0.128).
\begin{table}[ht]
  \centering\small
  \setlength{\tabcolsep}{8pt}
  \caption{Comparison of cell-token pooling strategies within 40$\times$ tiles. Performance is reported as mean per-gene PCC across the 5K panel; $\Delta$ denotes the absolute and relative change from mask-guided pooling.}
  \label{tab:cell_token}
  \begin{tabular}{lcc}
  \toprule
  Configuration & PCC (5K) & $\Delta$ \\
  \midrule
  \quad Cell mask (reference)                                   & \textbf{0.161 $\pm$ 0.016} & --- \\
  \quad Disc $r{=}5$\,\textmu m, Gaussian $\sigma{=}2.5$\,\textmu m & 0.154 $\pm$ 0.016      & $-$0.007 ($-$4.6\%) \\
  \quad Disc $r{=}5$\,\textmu m, unweighted mean                & 0.149 $\pm$ 0.015          & $-$0.012 ($-$7.3\%) \\
  \quad Class-token (baseline)                                  & 0.128 $\pm$ 0.013           & $-$0.035 ($-$20.2\%)\\
  \bottomrule
  \end{tabular}
\end{table}

\subsection{H\&E-derived centroids}
\label{app:he-derived-centroids}
\begin{figure}[htbp]
\centering
\begin{minipage}[c]{0.60\textwidth}
  \centering
  \captionof{table}{Agreement between CellViT++ nuclei centroids and Xenium platform-derived cell centroids. Values are median (interquartile range). Thresholds are cumulative: a pair counts if its distance is within the threshold in the nested one-to-one matching.}
\label{tab:cell-detection-match}
\small
\setlength{\tabcolsep}{3pt}
\resizebox{\linewidth}{!}{%
\begin{tabular}{lccc}
\toprule
Distance threshold & Precision & Recall & F1 \\
\midrule
$\leq 2~\mu$m & 0.758 (0.742--0.776) & 0.739 (0.706--0.764) & 0.748 (0.722--0.766) \\
$\leq 3~\mu$m & 0.853 (0.844--0.865) & 0.832 (0.809--0.851) & 0.841 (0.827--0.854) \\
$\leq 5~\mu$m & 0.901 (0.892--0.909) & 0.876 (0.855--0.890) & 0.887 (0.876--0.896) \\
\bottomrule
\end{tabular}}

\end{minipage}\hfill
\begin{minipage}[c]{0.35\textwidth}
  \centering
  \includegraphics[width=\linewidth]{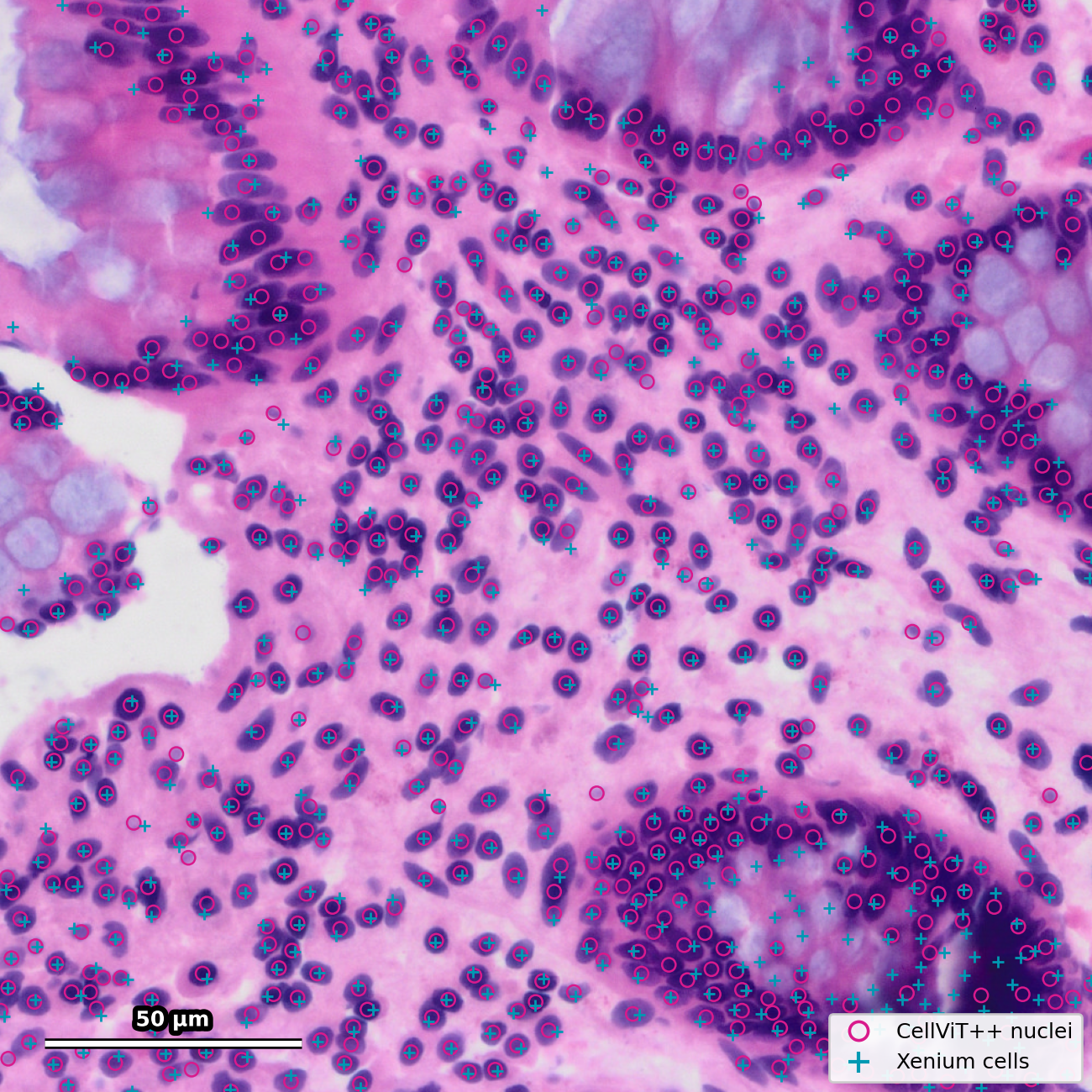}
  \captionof{figure}{CellViT++ nuclei centroids and Xenium cell centroids overlaid on a representative H\&E image.}
  \label{fig:cell-detection-example}
\end{minipage}
\end{figure}

To assess the concordance between cell coordinates derived from the Xenium platform and those derived from H\&E only, we generated nuclei centroids for all samples in the dataset using CellViT++~\cite{cellvit++} and the CellViT-SAM-H-x40-AMP backbone at 40$\times$ and 0.2125\,$\mu$m/pixel. This gave 2,373,383 nuclei across the 91 samples as compared to the 2,451,764 total cells from Xenium. Next, a one-to-one matching was performed in nested distance tiers of 2, 3, and 5\,$\mu$m, where at each tier, the optimal assignment among unmatched cells was computed with the Hungarian algorithm to maximize the number of pairs while minimizing the total distance between pairs. Matching resulted in a precision of 0.901 and recall of 0.876 using the conservative distance of between cell and nuclei centroids (Table~\ref{tab:cell-detection-match}, Figure~\ref{fig:cell-detection-example}).

\subsection{Input resolution ablation}
\label{app:input-res}
\begin{figure}[ht]
  \centering
  \includegraphics[width=0.95\linewidth]{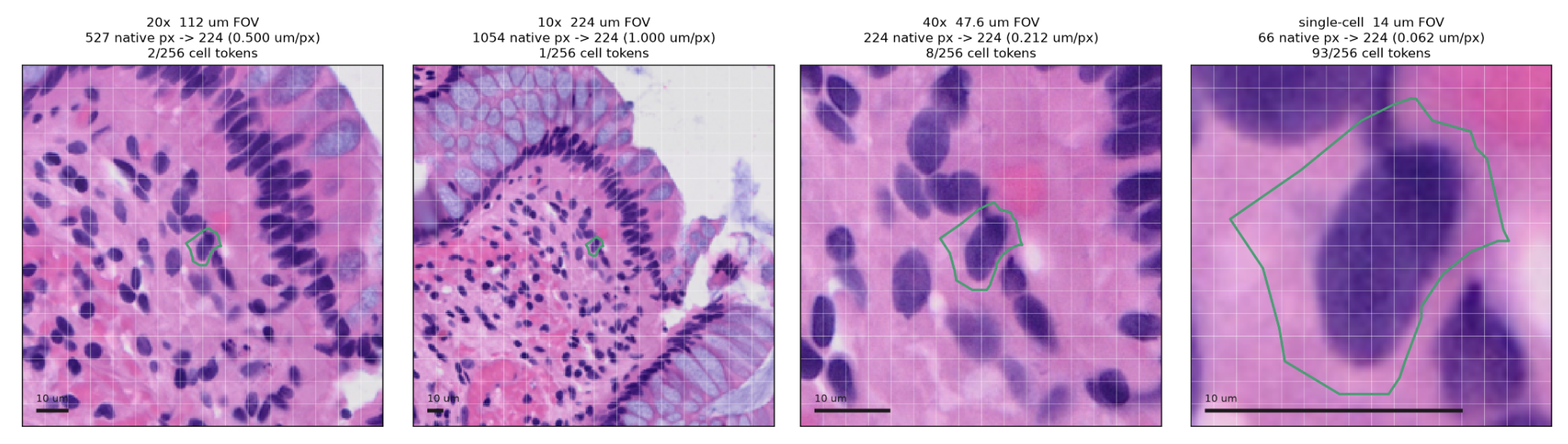}
  \caption{H\&E input configurations evaluated across spatial resolutions and fields of view. Each image is centered on the target cell, with its Xenium-derived boundary shown in green and patch-token boundaries overlaid as a grid.}
  \label{fig:fovs}
\end{figure}

We compared inputs spanning 10$\times$, 20$\times$, and 40$\times$ resolution, as well as 14$\times$14$\mu\mathrm{m}$ single-cell crops (Figure~\ref{fig:fovs}, Table~\ref{tab:input-resolution}). The 40$\times$, 47.6$\mu\mathrm{m} \times$47.6$\mu\mathrm{m}$ input performed best (PCC = 0.161). Lower-resolution inputs reduced performance by 5.6\% at 20$\times$ and 19.8\% at 10$\times$. Single-cell crops also reduced performance by 15.5\%, suggesting that prediction benefits from both fine cellular detail and surrounding tissue context.

 \begin{table}[ht]
    \centering\small
    \setlength{\tabcolsep}{5pt}
    \caption{Ablation of H\&E input resolution and field of view. Performance is reported as mean gene-wise PCC across the 5K panel; $\Delta$ denotes the absolute and relative change from 40$\times$.}
    \label{tab:input-resolution}
  \begin{tabular}{llllll}
    \toprule
    Input & Pixel size
          & FOV
          & Token width
          & Tokens/cell
          & PCC (5K) \quad \quad \quad \quad $\Delta$ \\
    & ($\mu$m/px) & ($\mu$m) & ($\mu$m) & median (mean) & \\
    \midrule
    $40\times$
      & 0.2125 & 47.6 & 2.98 & 4 (5.1)
      & \textbf{0.1611 $\pm$ 0.0158} \quad --- \\
    $20\times$
      & 0.5 & 112 & 7.00 & 1 (1.2)
      & 0.1521 $\pm$ 0.0125 \quad $-$5.6\% \\
    $10\times$
      & 1.0 & 224 & 14.0 & 1 (1.0)
      & 0.1292 $\pm$ 0.0107 \quad $-$19.8\% \\
    Single-cell crop
      & 0.2125 & 14.0 & 0.88 & 50 (56.8)
      & 0.1361 $\pm$ 0.0186 \quad $-$15.5\% \\
    \bottomrule
  \end{tabular}
\end{table}

\subsection{UMAP embedding representations} 
\label{app:umap}
\begin{figure}[ht]
  \centering
  \includegraphics[width=0.95\linewidth]{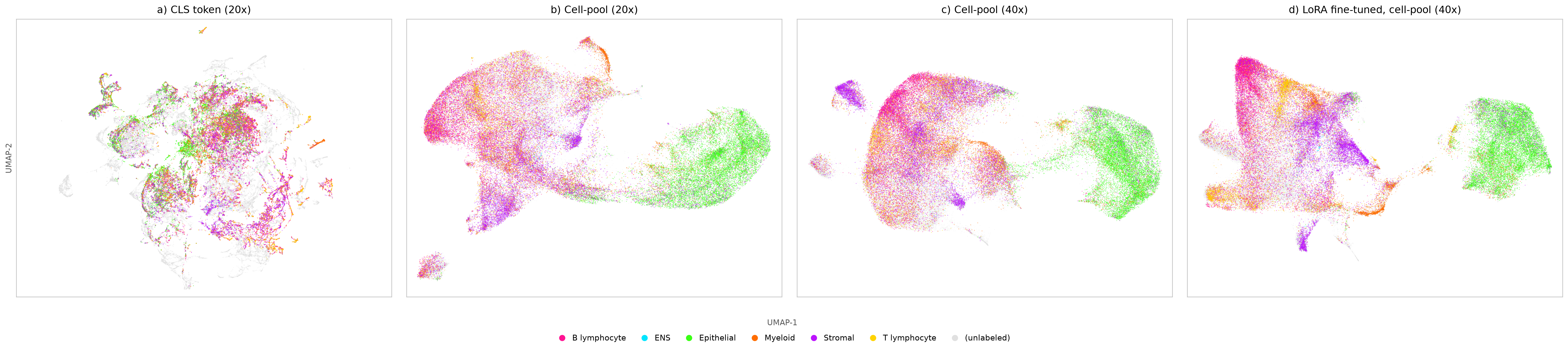}
  \caption{UMAP embeddings across adaptations, colored by cell type annotations.}
  \label{fig:umap}
\end{figure}

We sample 150,000 cells from one of the held out test folds, and plot a UMAP of the embeddings which were used for input to the prediction head. From left to right of Figure~\ref{fig:umap}, (a) shows the standard class-token embedding with 20$\times$ inputs, (b) shows the cell-token pooled embeddings also with 20$\times$ inputs, (c) shows the cell-token pooled embeddings with 40$\times$ inputs, and (d) shows the LoRA fine-tuned cell-token pooled embeddings.

\subsection{Epithelial gene predictions}
\label{app:epi-preds}
\begin{figure}[ht]
  \centering
  \includegraphics[width=0.95\linewidth]{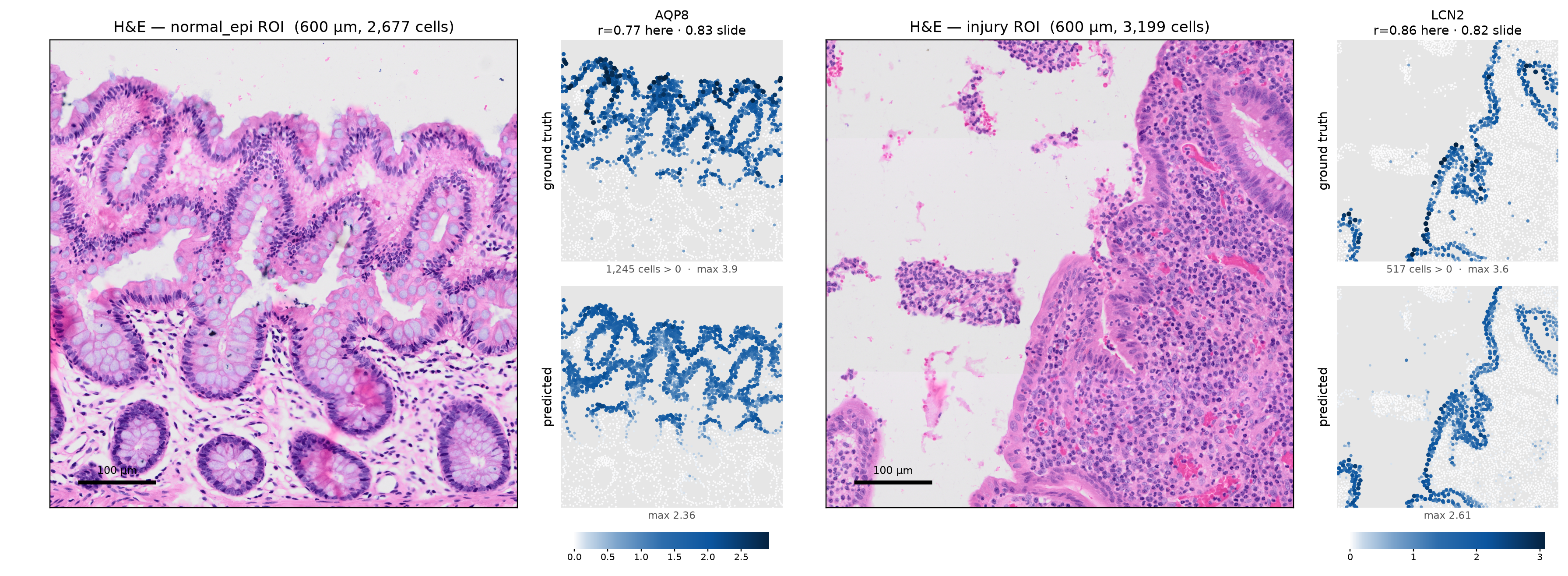}
  \caption{Predicted and observed single-cell expression of representative epithelial genes in a healthy region (left), and a diseased region (right).}
  \label{fig:epi_preds}
\end{figure}

Our model accurately predicts single-cell spatial gene expression of epithelial markers in both healthy and inflamed epithelial regions (Figure~\ref{fig:epi_preds}).

\subsection{Evaluation of spatially aggregated predictions}
\label{app:aggregated-preds}
\begin{figure}[ht]
  \centering
  \includegraphics[width=0.95\linewidth]{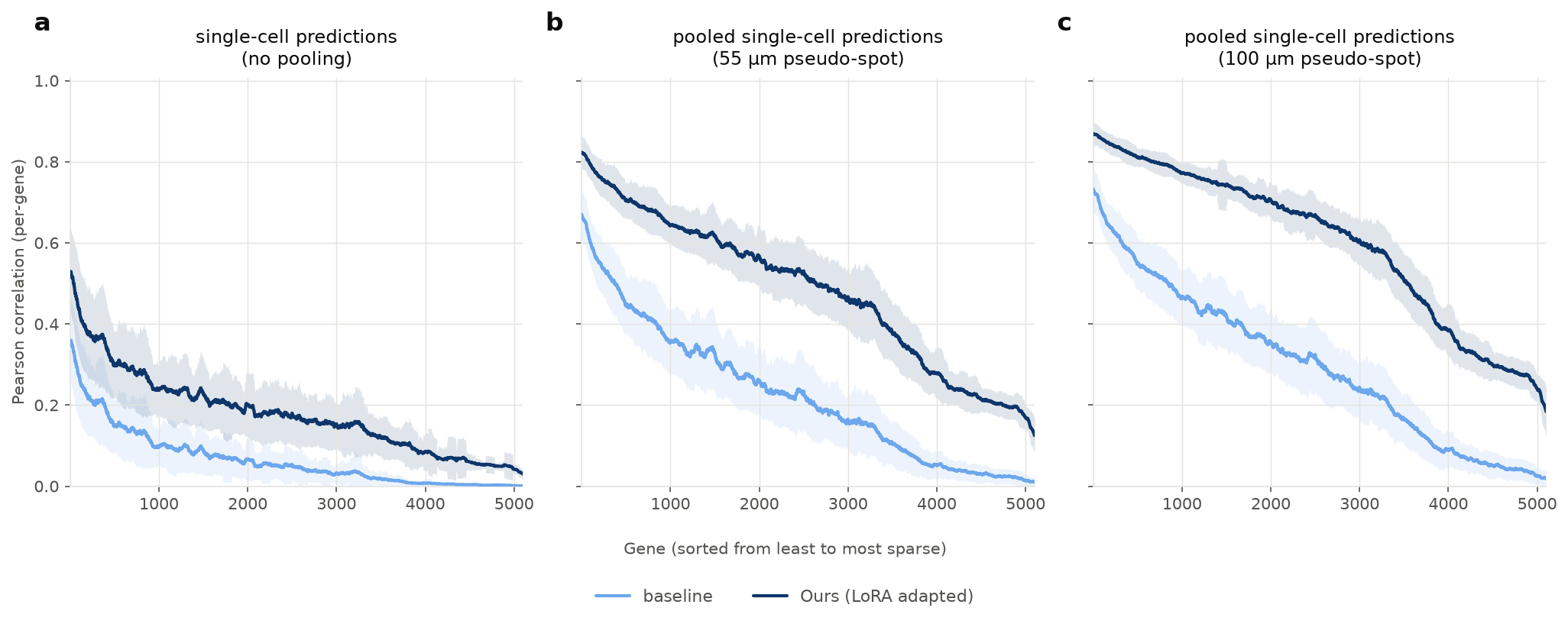}
  \caption{Per-gene PCC as a function of gene expression sparsity for the baseline linear probe vs. our adapted model. (a) Performance is evaluated at single-cell resolution and after aggregating predicted and measured single-cell expression into (b) 55$~\mu\mathrm{m}$ and (c) 100$~\mu\mathrm{m}$ pseudo-spots.}
  \label{fig:spot_vs_sc_combined}
\end{figure}

Our model maintained its advantage over the baseline across levels of gene expression sparsity, and the decline in performance for sparse genes became less pronounced after aggregation (Figure~\ref{fig:spot_vs_sc_combined}). This suggests that the adapted model captures morphology-associated transcriptional signals that are spatially coherent at both cellular and regional scales.

\subsection{Cell type annotation coverage}
\label{app:cell-type-coverage}
Cell type annotations were available for 41 of the 91 TMA cores, covering approximately 78\% of cells within those cores. Approximately 2.4 million TMA cells were scored for annotations, including cells with undetermined cell types. Of these cells, 31.4\% were assigned to one of the six classes: epithelial, stromal, B lymphocyte, myeloid, T lymphocyte, or enteric nervous system (ENS).

\subsection{Pathway enrichment analysis}
\label{app:pathway}
Using the top 500 most significantly improved genes, and the bottom 500 least improved genes through the adaptation changes to pathology FMs,  we sought to determine if the genes in both of these groups were enriched for specific biological programs and pathways. 

To this end, we used the g:Profiler~\cite{gprofiler} web tool to test both sets of genes for enrichment of specific pathways. For the top 500 most significantly improved genes (S1), we performed an ordered query with the default g:Profiler significance thresholds, and the REACTOME~\cite{reactome} database of pathways (Hsapiens (Human) - version: GRCh38.p14 - REAC – annotations: BioMart, classes: 2026-3-20). For the 500 least improved genes (S2), since order is not as straightforward to consider here, we performed an unordered query using the same REACTOME~\cite{reactome} pathway database, with the same default parameters for g:Profiler. Multiple testing correction was done using the g:SCS method~\cite{gprofiler} developed specifically for pathway analysis multiple hypothesis testing correction.

\begin{figure}[ht]
  \centering
  \includegraphics[width=0.95\linewidth]{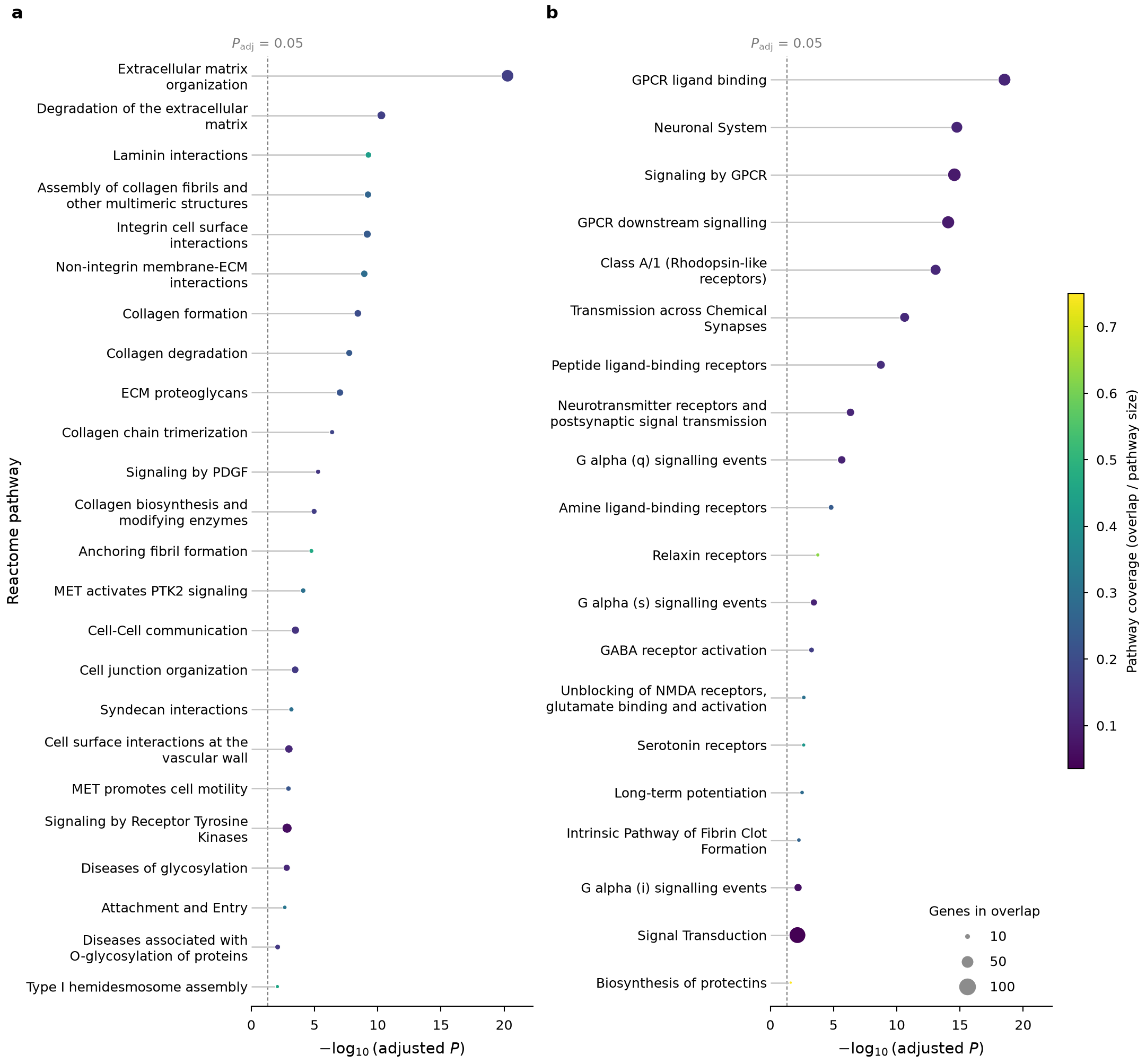}
  \caption{Pathway enrichment analysis was done using g:Profiler and the REACTOME database on the (a) 500 most improved genes and (b) 500 least improved genes.}
  \label{fig:pathways}
\end{figure}

All enriched pathways were plotted, sorted based on the adjusted p-values (significance), and overlap of genes in S1 and S2 in specific pathways/gene sets (Figure~\ref{fig:pathways}).

\end{document}